\documentclass[conference]{IEEEtran}
\usepackage{amsmath,amssymb,amsfonts}
\usepackage{algorithmic}
\usepackage{graphicx}
\usepackage{subcaption}
\usepackage{textcomp}
\usepackage{xcolor}
\usepackage{multirow}
\usepackage{pgfplots}
\usepackage{makecell}
\usepackage{booktabs}
\usepackage{romannum}

\DeclareMathOperator*{\argmax}{arg\,max}

\def\BibTeX{{\rm B\kern-.05em{\sc i\kern-.025em b}\kern-.08em
    T\kern-.1667em\lower.7ex\hbox{E}\kern-.125emX}}
\usepackage[style=ieee]{biblatex} 
\begin{document}

\title{Knowledge Graph Fusion for Language Model Fine-tuning}

\author{\IEEEauthorblockN{Nimesh Bhana}
\IEEEauthorblockA{\textit{Computer Science and Applied Mathematics} \\
\textit{University of Witwatersrand}\\
Johannesburg, South Africa \\
2371061@students.wits.ac.za}
\and
\IEEEauthorblockN{Terence L. van Zyl}
\IEEEauthorblockA{\textit{University of Johannesburg} \\
\textit{Institute for Intelligent Systems}\\
Johannesburg, South Africa \\
tvanzyl@gmail.com}
}

\maketitle

\begin{abstract}
Language Models such as BERT have grown in popularity due to their ability to be pre-trained and perform robustly on a wide range of Natural Language Processing tasks. Often seen as an evolution over traditional word embedding techniques, they can produce semantic representations of text, useful for tasks such as semantic similarity. However, state-of-the-art models often have high computational requirements and lack global context or domain knowledge which is required for complete language understanding. To address these limitations, we investigate the benefits of knowledge incorporation into the fine-tuning stages of BERT. An existing K-BERT model, which enriches sentences with triplets from a Knowledge Graph, is adapted for the English language and extended to inject contextually relevant information into sentences. As a side-effect, changes made to K-BERT for accommodating the English language also extend to other word-based languages. Experiments conducted indicate that injected knowledge introduces noise. We see statistically significant improvements for knowledge-driven tasks when this noise is minimised. We show evidence that, given the appropriate task, modest injection with relevant, high-quality knowledge is most performant.
\end{abstract}


\section{Introduction}
One major challenge in Natural Language Processing (NLP) is assessing the semantic similarity between pairs of text. Since natural language is versatile and ambiguous, it makes rule-based methods difficult to define~\cite{Marsi2013, standfordtyped, schuster-manning-2016-enhanced}. For instance, even though there may be a large discrepancy between the vocabulary used between two text sequences, they can have an equivalent meaning. This discrepancy is due to the polysemous property of most words and the use of synonyms, abbreviations, etc. The converse is also true: text sequences may have similar vocabulary but convey very different meanings, which is often caused by negations and, again, polysemy~\cite{homonymypoly}. Additionally, for domain-specific fields such as sports or medicine, knowledge of the underlying concepts is critical to computing accurate similarity measures~\cite{medicaldomain,oni2020comparative,manaka2022using}.

Humans typically rely on sources of information to fill the gaps in their knowledge. The same concept can be applied to machines. Sources of knowledge such as ontologies, thesauri, and knowledge bases have often been used to improve the performance of various NLP systems~\cite{majavu2008classification,RodriguezWordNet1185844,SemanticTextualSimilarityMethodsToolsAndAppsSurveyarticle}. However, successful integration between the NLP system and knowledge source often requires feature engineering. Furthermore, due to the structural inconsistencies between various knowledge sources, the integration is done in a tightly coupled manner.

The application of pre-trained Language Models for downstream tasks has produced state-of-the-art results in recent years~\cite{kooverjee2020inter,dlamini2019author}. However, the most performant models often contain billions of parameters, requiring expensive computational equipment. It is therefore vital to make such models as efficient as possible, which will also lead to less energy consumption.

Knowledge-Enhanced Models have demonstrated the power of leveraging external knowledge to improve the performance of pre-trained Language Models while retaining model size. The approach by Liu \textit{et al.}\ (2020) \cite{liu2019kbert} combines information from Knowledge Graphs with the ubiquitous BERT model by Delvin \textit{et al.}\ (2018) \cite{devlin-etal-2019-bert} as a means to complement input text with additional domain knowledge or contextual information. Although their results demonstrate benefits for knowledge-driven and domain-specific tasks, the selection mechanism for knowledge injection does not consider sentence context, which may lead to injecting irrelevant or unimportant information. Our study extends K-BERT by modifying the $K_\text{Query}$ mechanism to consider semantically important information and assessing its performance on both open-domain and domain-specific tasks. The utilisation of Wikidata as the Knowledge Graph is done loosely to ensure interchangeability with other knowledge sources. Using ablation studies, we examine the type of knowledge most beneficial to the fine-tuning process and associated limitations. Lastly, since K-BERT was initially developed for the Chinese language, there is a lack of research indicating the level of versatility in the underlying architecture when applied to other languages.

To this end, the article presents the results of our investigation, which show that:
\begin{enumerate}
    \item K-BERT can be successfully adapted to the English domain and possibly other languages.
    \item A modified $K_\text{Query}$ to only inject semantically related information can be beneficial.
    \item Inclusion of external knowledge introduces noise. 
    \item In the absence of noise, external knowledge injection is beneficial to knowledge-driven tasks.
\end{enumerate}


\section{Related Work}
\label{RelatedWork}
The ever-growing corpora of publicly available textual data have accelerated the growth of machine learning in NLP. Neural models can now be pre-trained with general domain information and used or fine-tuned further for downstream tasks. Word2vec \cite{mikolov2013efficient} being one of the most popular pre-trained word embeddings. 

Sinora \textit{et al.}\ (2019)~\cite{SINOARA2019955} use knowledge bases and pre-trained embeddings to construct embedded representations of documents for text classification. Their knowledge base, BabelNet, is used to disambiguate word senses, allowing them to create more accurate vector representations of documents. NASARI vector representations and word2vec embeddings are used to compute an average synset vector representing documents. While they were able to represent word senses in a more meaningful manner, word2vec did perform better in specific settings. Disambiguating the sense of each word was not always successful. Results could potentially be improved if this issue is resolved.

To \textit{semantify} input text, Pilehvar \textit{et al.}\ (2017) \cite{towardsaseemlessintegration} integrates sense-level knowledge into a CNN text classifier by disambiguating input text using information from a WordNet semantic network. This aim is to disambiguate text before being fed into a system. They highlight that simple input disambiguation can bring about a performance gain for longer texts. Since their disambiguation algorithm relies on constructing a graph from the input text, short texts do not provide sufficient connections to function effectively, producing lower performance. 

More modern research investigates the concept of incorporating external knowledge into pre-trained Language Models. Nguyen \textit{et al.}\ (2019) \cite{NGUYEN2019104842} can achieve performance that slightly outperformed BERT. They exploit the similarity of word contexts built by word embeddings and semantic relatedness between concepts based on external knowledge sources such as WordNet and Wiktionary. Using their approach, they can more accurately assess the similarity between two short texts. However, their approach is supervised and requires feature engineering.

Lu. et al.\ (2020) \cite{lu2020vgcnbert} combine the capability of BERT with Vocabulary Graph Convolutional Network (VGCN). VGCN convolves directly on a vocabulary graph to induce embeddings based on node properties and neighbourhoods. Their experiments show that a combination of the local information from BERT and global information from VGCN outperforms BERT and VGCN alone. Their vocabulary graph is constructed using normalised point-wise mutual information (NPMI) and word co-occurrences with documents. Conclusion remarks suggest lexical resources such as WordNet, or combinations of such, may contain valuable connections that NPMI cannot cover and should be explored.

Alternative approaches involve amalgamating a separate word embedding with its associated node embedding from a Knowledge Graph. Construction of embeddings from Knowledge Graphs utilises Graph Convolutional Networks to produce embeddings for every entity or node. Since these embeddings are produced separately, their vector spaces are inconsistent with each other. The K-BERT model by Liu \textit{et al.}\ (2020) \cite{liu2019kbert} avoids this by injecting information before the embedding process. They inject knowledge into sentences using Chinese Knowledge Graphs (CN-DBpedia, HowNet, MedicalKG). These knowledge-rich sentences are then passed through to a BERT-based model to learn more meaningful sentence representations. This work extends K-BERT by adapting the model to the English domain.

\section{Method: English adapted K-BERT}
\label{Method}

\subsection{Knowledge Graph}
\label{KnowledgeGraph}
This work employs Wikidata as the primary knowledge source to retrieve information. As with most knowledge graphs, it can be stored in a triplet format, i.e. (subject, predicate, object). An example statement could be (Michelle Obama, wifeOf, Brack Obama), where the subject and object, in this case, are entities and the predicate is a property. Entities extend to a range of other data items such as locations, celebrities, concepts, and even simple terms such as "bank". Each data item contains various statements describing itself and any relationships it may have\footnote{\url{https://www.wikidata.org/wiki/Help:Items} (2016)}. An advantage of using Wikidata is that data items are inherently unambiguous. The following subsections will show how the exact property can be exploited for knowledge injection.

\subsubsection{Data Preprocessing}
Since Wikidata contains almost one hundred million data items\footnote{\url{https://www.wikidata.org/wiki/Wikidata:Main_Page} (2022)}, integration of the entire Knowledge Graph is expensive and infeasible. We reduce this size by only considering English data items in domains: business, sports, humans, cities, and countries. Furthermore, only properties belonging to the set \{label, alias, description, subclass of, instance of\} are used. These properties were selected because they maximise descriptive detail while minimising storage requirements. An additional benefit is that other Knowledge Graphs are likely to contain similar information since they are general properties. It is easy to swap out Wikidata for an alternative Knowledge Graph without drastically affecting performance. 

\subsection{Term-based Sentence Tree}
Since Chinese is logographic, K-BERT was designed to work on a character level. Model inputs such as the sentence tree and visible matrix must be adapted to accommodate the alphabetic English language on a word level. However, a further extension to the multi-word-level or \textit{term}-level is also necessary as there are cases where entities may span across two or more words in a sentence (e.g. name of a person or place). Knowledge injection is done per \textbf{group} of tokens instead of a single token. Specifically, given an input sentence $s = [w_0,w_1,\ldots,w_n]$ comprised of tokens $w_i$ in vocabulary $V$, enclose contiguous related tokens $w_o$ to $w_q$ together in a group ${ }$. It is not required that a group consists of multiple tokens. Therefore, we can have a particular group ${w_j}$, where $w_j$ has no other contiguously related tokens. Knowledge is injected per group of related tokens to produce an output sentence tree $t$:
\begin{multline}
      t = \Big[\{w_0\}, \{w_1\},\ldots, \Big.\\ 
    \{w_o, \ldots, w_q\}\big[(r_{op0}, w_{op0}), \ldots, (r_{opk}, w_{opk})\big], \ldots,\\
    \Big. \{w_n\}\Big]
\end{multline}
where $w \in V$ is the set of entities in the Knowledge Graph $\mathrm{K}$, $r \in V$ is the set of relations/properties in the Knowledge Graph, and $k$ represents the number of triplets inserted. Braces enclose related tokens, and their knowledge, if any, is inserted directly afterwards. Construction of the visible matrix and positional encoding can be done similarly. However, an additional "unrolling" step is required to associate positions per token. As the last step, the visible matrix ensures token groups and their knowledge all "see" each other by appointing a value of 1 in their cell. This optimisation alleviates duplicating the same knowledge for each token in the group, thereby minimising sequence lengths provided to the model.

\subsection{Contextualised Knowledge Injection}
In this section, we highlight the approach taken to inject knowledge into sentences. For an input sentence $s$, perform named entity recognition to extract entities such as names of people, places, sports teams, etc. We accomplish this with a small pre-trained model from spaCy\footnote{https://spacy.io/} based on a transition-based parser \cite{lample-etal-2016-neural, honnibal_2013} and an "embed, encode, attend, predict" framework \cite{honnibal_2016}. After that, the extracted entities are queried from the Knowledge Graph to retrieve a list of triplets. This processing is done by $K_\text{Query}$:
\begin{equation}
    E = K_\text{Query}(s, \mathrm{K}) 
\end{equation}
where $E$ is a collection $[(w_i, r_{i0}, w_{i0}), \ldots, (w_i, r_{ik}, w_{ik})]$ of triplets. The function, $K_\text{Inject}$, then injects $E$ into the correct position and generates the sentence tree $t$ as:
\begin{equation}
    t = K_\text{Inject}(s, E)
\end{equation}

However, additional processing to find the most relevant triplets is done before injection occurs. Hence, instead of injecting the first $n$ entities as done by Liu \textit{et al.}\ (2020) \cite{liu2019kbert}, we use a pre-trained Transformer model to determine which retrieved entities from $E$ are most relevant to the current sentence $s$. This injection is done by concatenating each entity and their related properties into a single text sequence $\operatorname{seq}_i$. The pre-trained Transformer model $T$ then generates contextualised embeddings $\operatorname{emb}_i$ for each sequence $\operatorname{seq}_i$ using:
\begin{equation}
    \operatorname{emb}_i = T(\operatorname{seq}_i)
\end{equation}

Contextualised embedding $\operatorname{emb}_s$ is generated for sentence $s$ in the same manner. Similarity between embeddings is computed using cosine similarity metric $\left\|\cdot, \cdot\right\|_\text{cos}$. The entity corresponding to the most similar embedding is selected to be injected using:
\begin{equation}
    \operatorname{max\_pos} = \argmax_{i \in I}\left(\left\|\operatorname{emb}_i, \operatorname{emb}_s\right\|_\text{cos}\right)
\end{equation}
where $I$ is the set of distinct entities.In order to limit the amount of noise injected, we introduce a threshold parameter such that information is only injected if $\left\|\operatorname{emb}_i, \operatorname{emb}_s\right\|_\text{cos} > \text{threshold}$. 

\subsection{Corrections \& Optimisations}
\subsubsection{Sequence Truncation}
\begin{figure}[htbp]
\centering
\begin{subfigure}{.45\textwidth}
  \centering
  \includegraphics[width=.9\linewidth]{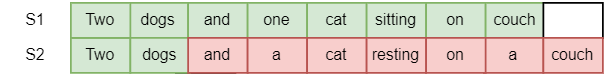}
  \caption{Previous truncation method}
  \label{fig:truncationa}
\end{subfigure}
\begin{subfigure}{.45\textwidth}
  \centering
  \includegraphics[width=.9\linewidth]{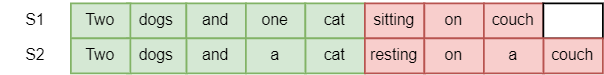}
  \caption{Truncating sentences equally}
  \label{fig:truncationb}
\end{subfigure}
\caption{Comparison of truncation methods $\operatorname{max\_length}=10$. Words highlighted in red are truncated, green words are kept.}
\label{fig:truncationmethods}
\end{figure}

Truncation has been modified to be more "equal" when sentence pairs are fed into K-BERT. Given a $\operatorname{max\_length}$, the number of tokens in S1 and S2 is each limited to size $\operatorname{max\_length}/2$. Unoccupied positions are assigned as leftover and given to the remaining sentence that requires it. The difference is illustrated in Fig.~\ref{fig:truncationmethods} for sentences S1 and S2.

\subsubsection{Memory Usage}
Dimension of the visible matrix is $(\operatorname{max\_length} \times \operatorname{max\_length})$. Implying that memory usage grows exponentially as sequence length increases. Such growth can become infeasible for large datasets containing upward of $100,000$ samples such as AG News Subset \cite{AGNEWS}. Since the visible matrix is symmetrical, we optimise memory usage by intermediately storing non-duplicate entries from the visible matrix into a vector of size $N (N + 1)/2$. When the matrix is required to be fed into K-BERT, we convert the vector back into a symmetrical matrix - done per batch. Although this does require additional computation, the trade-off is worthwhile - especially if memory capacity is a bottleneck. Lastly, since the visible matrix strictly contains binary values ("visible", "invisible"), a final memory optimisation explicitly restricts values to one byte each. With the described optimisations, the visible matrix memory consumption is reduced by a factor of four \emph{at minimum}.

\section{Evaluation}
\label{Experiment}
\subsection{Datasets}
This work focuses on the task of semantic similarity. We consider two widely used and credible datasets. The Semantic Textual Similarity Benchmark (STS-B) \cite{stsb} was used in the public domain. STS-B contains pairs of sentences that human judges have manually annotated according to how similar each sentence pair is. Each pair is assigned an averaged score between $0$ and $5$, with higher scores implying higher similarity.

Since domain-specific datasets equivalent to STS-B are expensive to construct, we consider a text classification dataset ag\_news\_subset \cite{AGNEWS}. This dataset contains extracts from news articles, and the task is to classify articles into the correct category from set \{World, Sci/Tech, Sports, Business\}. Since this task requires domain-specific knowledge in the fields mentioned above to classify articles correctly, it is appropriate to assess model performance for domain-specific problems.

\subsection{Experimental Setup}
This section describes the experimental setup for comparing BERT to K-BERT. Hyperparameters and additional configurations were kept consistent across both models. Specific seeds were used for all runs to ensure result reproducibility. Each model was fine-tuned for ten epochs. The best-performing model on the validation set was saved and used to evaluate the test set. Batch sizes of $16$ and $32$ were used for STS-B and ag\_news\_subset, respectively. Being cognizant of the catastrophic forgetting problem \cite{catastrophicforgetting, catastrophicbert}, only modest learning rates were considered as hyperparameters ($2e^{-5}$ to $5e^{-5}$). Based on the validation set, the best performing learning rates were $4e^{-5}$ and $5e^{-5}$ for STSb and ag\_news\_subset, respectively.

Hyperparameters were tuned according to the best performance on the validation set. The best threshold parameter, described in section \ref{Method}, was $0.5$ for STS-B and $0.6$ for ag\_news\_subset. Our available resources necessitated a balance between computational requirements and performance. Therefore, a maximum sequence length of $128$ for ag\_news\_subset and 256 for STS-B was set. Additionally, the pre-trained $\text{BERT}_{\text{BASE}}$ was used instead of the better performing, yet computationally demanding, $\text{BERT}_{\text{LARGE}}$.

Adam optimiser with weight decay and scheduler parameters remained unchanged from the work by Liu \textit{et al.}\ (2020) \cite{liu2019kbert} and Devlin \textit{et al.}\ (2018) \cite{devlin-etal-2019-bert}. That is, Adam with $L_2$ weight de

\begin{table}[ht]
    \caption{Evaluation Datasets}
    \label{tab:evaluationdatasets}
    \centering
    \begin{tabular}{l|rrr}
        \toprule
         \textbf{Dataset} & \textbf{Train} & \textbf{Validation} & \textbf{Test} \\ \bottomrule\toprule
         STS-B & 5,749 & 1,500 & 1,379 \\
         ag\_news\_subset & 110,400 & 9,600 & 7,600 \\ \bottomrule
    \end{tabular}
\end{table}

\subsection{Experiment}
An extended comparison between the two models has been made to explore the benefits and deficiencies of knowledge incorporation into BERT. In order to limit biases and highlight statistical significance, results are averaged over ten runs. Each of these runs uses consecutive seeds ($8$ to $17$) to ensure the reproducibility of results.

We aim to attain insight into \romannum{1}) what type of knowledge is most beneficial and \romannum{2}) at what point does knowledge incorporation become detrimental to performance. The following experiments are performed to gain more insight:
\begin{enumerate}
    \item Knowledge Ablation: The type of knowledge injected (as described in Section \ref{KnowledgeGraph}) can be categorised into aliases, categorical, and descriptive information. To identify what information is most beneficial, we exclude each type from being injected into K-BERT and rerun experiments.
    \item Knowledge-Gating: Since K-BERT requires sequences to be truncated passed a maximum sequence length, injecting additional information can lead to important information from the original sentence being truncated. To see the extent to which this is true, we perform Knowledge-Gating. That is, only inject information into sentences when sequence lengths are below the maximum sequence length - thereby avoiding truncation.
    \item Manual Knowledge Injection: A manual selection of knowledge is performed to identify deficiencies in the automated similarity-based approach presented in $K_{Inject}$ for relevant knowledge fusion. The manual approach is expected to contain minimum noise since humans can match relevant knowledge to sentence context to a much higher degree. Manual Knowledge Injection is done for all dataset splits as described in detail below.
\end{enumerate}

\subsubsection{Manual Knowledge Injection}
More specifically, the systematic approach for performing Manual Knowledge Injection requires manually matching extracted entities $E$ to the context of sentence $s$. When sentence $s$ lacks context, description information is typically selected to be injected. In situations where descriptive information would introduce noise or be inappropriate, we default to categorical or alias information. While most of this process is subjective, the primary objective of only injecting beneficial information for the problem at hand is maintained throughout. For example, the occurrence of an entity "Manchester United" in ag\_news\_subset will likely be associated with external knowledge "football club" - \textbf{if} the terms do not already exist in the original sentence $s$. 
\section{Results}
\label{Results}

Table~\ref{tab:knowledgeablationsts} shows the Knowledge Ablation results on the STS-B dataset. We report total Mean Squared Error (MSE) loss and Spearman correlation. The correlations are between the predicated and actual scores. Combining knowledge categories from set {ALIAS, CAT, DESC} were removed from K-BERT, and the results are presented here. Abbreviated knowledge categories consecutively correspond to {aliases, categorical information, descriptive information}. Additionally, a model with minimal noise, $\text{K-BERT}_{\text{MANUAL}}$, indicates that knowledge selection is made manually by humans instead of the similarity-based approach of K-BERT.

For ag\_news\_subset, we report classification accuracies for BERT and all K-BERT ablation variations in Table~\ref{tab:knowledgeablationcls}. Compared to STS-B, sequence length limitations is a more stringent factor for ag\_news\_subset due to its much longer paragraphs of text. The $128$ sequence length limitation we impose on it has further exacerbated this factor. Knowledge-Gating experiments have been performed to investigate the effect of retaining maximal information from the original sentences. The results of this are shown in column Knowledge-Gating=on. 

\begin{table}[ht]
	\caption{Knowledge Ablation results on STS-B dataset for mean ($\mu$-validation / $\mu$-test $\pm \sigma$) scores taken across ten runs. (p-values) indicate that none of the results are a statically significant improvement on the baseline: BERT.}\label{tab:knowledgeablationsts}
	\centering
	\resizebox{\columnwidth}{!}{%
	\begin{tabular}{l|rr|rr}
		\toprule
		\textbf{Model} & \textbf{MSE Loss} & p & \textbf{Spearman} & p \\
		\bottomrule\toprule
		BERT            & 1.662 / 2.014 $\pm 0.054$ &     & 89.67 / 85.73 $\pm 0.49$ & \\
	    K-BERT          & 1.675 / 2.028 $\pm 0.063$ & 0.6 & 89.58 / 85.66 $\pm 0.56$ & 0.5 \\ \midrule
        \textbf{After Removing} & & & \\
        \, ALIAS        & 1.655 / 2.025 $\pm 0.055$ & 0.6 & 89.65 / 85.71 $\pm 0.47$ & 0.5 \\
        \, CAT          & 1.655 / 2.018 $\pm 0.059$ & 0.5 & 89.63 / 85.64 $\pm 0.54$ & 0.6 \\
        \, DESC         & 1.667 / 2.019 $\pm 0.056$ & 0.5 & 89.60 / 85.76 $\pm 0.49$ & 0.4 \\
        \, CAT + DESC   & 1.667 / 2.009 $\pm 0.068$ & 0.4 & 89.54 / 85.80 $\pm 0.57$ & 0.3 \\
        \, ALIAS + DESC & 1.664 / 2.018 $\pm 0.059$ & 0.5 & 89.59 / 85.80 $\pm 0.44$ & 0.3 \\
        \, ALIAS + CAT  & 1.653 / 2.009 $\pm 0.058$ & 0.4 & 89.59 / 85.77 $\pm 0.50$ & 0.4 \\
    \midrule
        $\text{K-BERT}_{\text{MANUAL}}$ & 1.657 / 2.024 $\pm 0.059$ & 0.6 & 89.63 / 85.65 $\pm 0.5714$ & 0.6 \\
		\bottomrule
	\end{tabular}}
\end{table}

\begin{table*}[htb!]
    \caption{Knowledge Ablation and Knowledge-Gating results on ag\_news\_subset dataset. We report mean ($\mu$-validation / $\mu$-test $\pm \sigma$) accuracies taken across ten runs. (p-values) in bold indicate that none of the results are a statically significant improvement on the baseline: BERT}\label{tab:knowledgeablationcls}
    \centering
    \begin{tabular}{l|rr|rr}
        \toprule
        \multirow{2}{*}{\textbf{Model}} & \multicolumn{4}{c}{\textbf{Knowledge-Gating}} \\
        & \makecell[c]{\textbf{Off}} & (p-value) &  \makecell[c]{\textbf{On}} & (p-value) \\
        \bottomrule\toprule
        BERT            & 94.55 / 94.60 $\pm 0.1947$ & & &\\
        K-BERT          & 94.55 / 94.65 $\pm 0.2805$ & (0.3518) & 94.58 / 94.69 $\pm 0.1269$ & (0.1328)\\ \midrule
        After Removing & & \\
        \, ALIAS        & 94.60 / 94.60 $\pm 0.2272$ & (0.5039) & 94.52 / 94.67 $\pm 0.1138$ & (0.1854) \\
        \, CAT          & 94.53 / 94.61 $\pm 0.1144$ & (0.4791) & 94.53 / 94.67 $\pm 0.1474$ & (0.2072) \\
        \, DESC         & 94.54 / 94.63 $\pm 0.1409$ & (0.3517) & 94.62 / 94.63 $\pm 0.2274$ & (0.3911) \\
        \, CAT + DESC   & 94.61 / 94.59 $\pm 0.2071$ & (0.5538) & 94.55 / 94.60 $\pm 0.2070$ & (0.5290) \\
        \, ALIAS + DESC & 94.53 / 94.59 $\pm 0.1378$ & (0.5640) & 94.61 / 94.67 $\pm 0.0989$ & (0.1745) \\
        \, ALIAS + CAT  & 94.57 / 94.63 $\pm 0.1585$ & (0.3665) & 94.53 / 94.54 $\pm 0.1113$ & (0.7948) \\
        \midrule
        $\text{K-BERT}_{\text{MANUAL}}$ & \textbf{95.16 / 95.24} $\pm 0.1361$ & \textbf{(0.0000)} & \textbf{95.15 / 95.23} $\pm 0.1824$ & \textbf{(0.0000)} \\
        \bottomrule
    \end{tabular}
\end{table*}

\section{Analysis}\label{Analysis}
To identify the significance of the results, we perform one-tail Student t-tests between test results from BERT and K-BERT. Results were recorded after repeating the exact Experiment ten times using different seeds. These are the same results used to produce the averages reported in section \ref{Results}. Independent sample t-tests are performed between BERT results and K-BERT Knowledge Ablations. We begin this section with an analysis of the results of the STS-B dataset. Analysis of ag\_news\_subset results is done after that.

\subsection{STS-B Analysis}
Inspecting the K-BERT model results in Table~\ref{tab:knowledgeablationsts}, we see an overall reduction in performance with the addition of knowledge in the BERT model. They are implying that the knowledge is introducing noise. It is expected that reducing this noise will improve performance. The Knowledge Ablation experiments described in section~\ref{Experiment} look to evaluate if this noise is the cause of the decline in performance.

After performing Knowledge Ablation, overall performance improves to a marginal degree compared to the standard K-BERT. Descriptive information tends to be the most beneficial \textit{type} of knowledge to inject, with categorical information also providing some benefit. These insights can be identified by inspecting the changes in metrics when descriptive and categorical information is removed from the model. The ALIAS + CAT variant produced the lowest MSE loss, while on average, ALIAS + DESC had the highest correlation coefficients. 

However, overall results for this dataset indicate that noise exists in all knowledge types. Moreover, a manual injection of knowledge could not improve performance at all. Similar to other research \cite{dropredundant} which injects knowledge in pre-trained language models for the GLUE benchmark dataset \cite{GLUE}, it can be concluded that the inclusion of knowledge for this problem is not beneficial. This result is further substantiated by insignificant t-test results in Table~\ref{table:ststtest}.

\subsubsection{STS-B Sentence Pair Ablation Study}
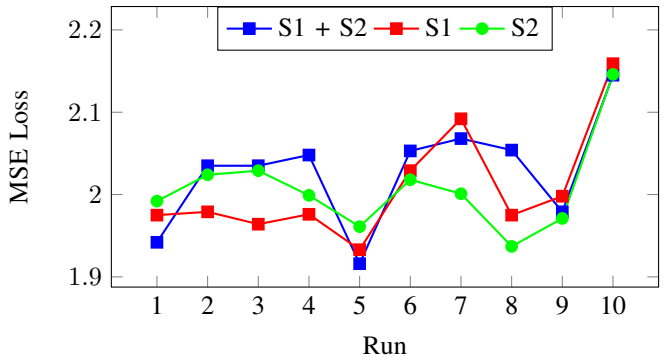
\begin{figure}
	\centering
	\begin{tikzpicture}
	\begin{axis}[
	width=\columnwidth,
	height=.6\columnwidth,
	legend style={at={(0.5,1)},
		anchor=north,legend columns=-1},
	symbolic x coords={1, 2, 3, 4, 5, 6, 7, 8, 9, 10},
	xticklabels={1, 2, 3, 4, 5, 6, 7, 8, 9, 10},
	enlargelimits = true,
	xtick=data,
		ymax=2.2,
	ylabel={MSE Loss},
	xlabel={Run}
	]
	\addplot[blue,thick,mark=square*]  table[x=epoch,y=kbert]{\stsablationlossdatatest};
	\addplot[red,thick,mark=square*]  table[x=epoch,y=seq0]{\stsablationlossdatatest};
	\addplot[green,thick,mark=otimes*] table[x=epoch,y=seq1]{\stsablationlossdatatest};
	\legend{S1 + S2, S1, S2}
	\end{axis}
	\end{tikzpicture}
    \caption{Test MSE loss on best performing model per run.}
	\label{stslossablationplottest}
\end{figure}

The STS-B dataset contains pairs of text which may be similar. Naturally, it is possible for both sentences to contain the exact words and therefore have the same information injected into them - possibly biasing sentences to have a higher similarity by nature of the method. An ablation study was performed to investigate if such a bias does exist. Two additional experiments were conducted. The first exclusively injects information for sentence S1. While the second exclusively injects information for sentence S2. The experiment results were compared to K-BERT, which has knowledge injected into both sentences (S1 + S2). As shown in Table~\ref{tab:stsablation}, which contains the average test results of the best performing model across ten runs, no significant bias exists. Additional evidence can be seen in Fig.~\ref{stslossablationplottest}, where the loss of K-BERT is generally higher across the ten runs instead of lower - which would have indicated a bias. 
\begin{table}[ht]
    \caption{Comparison of test results when exclusively injecting knowledge for sequences S1, S2 and both S1 + S2.}
	\label{tab:stsablation}
	\centering
    \begin{tabular}{l|rrr}
    	\toprule
    	\textbf{Metric} & \textbf{S1 + S2} & \textbf{S1} & \textbf{S2} \\
    	\bottomrule\toprule
    	MSE Loss & 2.028 & 2.008 & 2.008 \\ 
    	Spearman & 85.66 & 85.83 & 85.78 \\ \bottomrule
	\end{tabular}
\end{table}

\pgfplotstableread[row sep=\\,col sep=&]{
epoch & kbert & seq0 & seq1 \\
1 & 1.942 & 1.975 & 1.992 \\
2 & 2.035 & 1.979 & 2.024 \\
3 & 2.035 & 1.964 & 2.029 \\
4 & 2.048 & 1.976 & 1.999 \\
5 & 1.916 & 1.933 & 1.961 \\
6 & 2.053 & 2.029 & 2.018 \\
7 & 2.068 & 2.092 & 2.001 \\
8 & 2.054 & 1.975 & 1.937 \\
9 & 1.979 & 1.998 & 1.971 \\
10 & 2.145 & 2.159 & 2.146 \\
}\stsablationlossdatatest

\subsection{AG News Subset Analysis}
In contrast to STS-B, the ag\_news\_subset dataset appears to benefit more from the inclusion of knowledge. K-BERT has an improved average test accuracy compared to BERT, as seen in Table~\ref{tab:knowledgeablationcls}. Removing noise through Knowledge Ablation does not benefit the test performance compared to K-BERT. However, these test accuracies are generally improved with Knowledge-Gating=on, when truncation is avoided. As we have seen with STS-B, all knowledge types introduce noise. Descriptive and categorical information still appears to be the most beneficial for this classification problem, with aliases providing a more significant benefit than in STS-B. 

Although the reported average accuracies indicate gradual performance improvements, the t-test results in Table~\ref{table:clsttest} conclude that the difference is statistically insignificant. Further noise removal through $\text{K-BERT}_{\text{MANUAL}}$, however, produces the best results compared to every K-BERT variation as well as the original BERT model. When Knowledge-Gating=on, performance is negatively affected to a very marginal degree. This result highlights that, in the absence of noise, additional knowledge provides contextual information that can be more beneficial for this problem than the current information in the original sentences. Since the extracted entities for this problem, such as \textit{Intel, Microsoft, Apple, Toyota, Yankees} occur relatively frequently within the dataset, the associated additional knowledge enforces computed embeddings to be aggregated into more well-defined clusters. With a $p-value=0.0$, $\text{K-BERT}_{\text{MANUAL}}$ has shown a statistically significant benefit for the inclusion of knowledge. 

In conclusion, the fusion of knowledge from the Wikidata Knowledge Graph has potential benefits. When inspecting Table~\ref{tab:knowledgeablationcls}, K-BERT with Knowledge-Gating=on produces better results and has more significant t-test results for most models. However, all autonomous approaches and knowledge types introduce some noise which causes a decline in performance. When minimising the amount of noise, the STS-B dataset exhibits no benefit from the fusion of knowledge, while the ag\_news\_subset dataset produces a $0.7\%$ improvement over BERT. Despite the potential benefits of incorporating external knowledge, achieving successful integration with an appropriate problem autonomously is non-trivial. 

\section{Conclusion and Future Work}
\label{Conclusion}
This article presents an extended K-BERT model, adapted to accommodate the English language, along with a new Wikidata Knowledge Graph. The sentence tree and visible matrix have been changed to work on a term level instead of the character level meant for Chinese. Sequence truncation has been made to be done more equally between sentence pairs. Additional optimisations permit larger datasets to use less memory.

Knowledge Ablation studies indicate that while knowledge injection benefits performance on average, it introduces noise which hurts performance. Reducing the amount of noise leads to more significant results on the ag\_news\_subset dataset. However, the same does not hold for STS-B. As corroborated in \cite{dropredundant, ERNIE}, knowledge-enhanced models such as ERNIE are unstable on smaller datasets such as STS-B and ultimately perform worse. The drop in performance can be attributed to STS-B benefits more from improved linguistic representation than structured facts. Therefore, we conclude that given the appropriate problem, injecting knowledge sparingly with relevant, high-quality information is preferable. 

Since the contextual mechanism in $K_\text{Query}$ was intended to work on the general text, replacing the Knowledge Graph with free text is possible. Additionally, the multilingual capability of Wikidata allows K-BERT to accommodate various other languages, given the appropriate pre-trained Language Model. Future work should explore advanced contextual mechanisms which consider additional factors other than similarity. Possibilities include a pre-disambiguation step or machine learning models to decide on the entity to inject.


\printbibliography 

\section*{Appendix}
\begin{table}[ht]
    \caption{Student t-test results between metrics of BERT \& K-BERT Models evaluated on STS-B test dataset}
	\label{table:ststtest}
	\centering
	\begin{tabular}{l|rr|rr}
		\toprule
        \multirow{3}{*}{\textbf{Model}} & \multicolumn{4}{c}{\textbf{Metric}} \\
        & \multicolumn{2}{c}{\textbf{MSE Loss}} & \multicolumn{2}{c}{\textbf{Spearman}} \\
        & \textbf{t-statistic} & \textbf{p-value} & \textbf{t-statistic} & \textbf{p-value} \\
        \bottomrule\toprule
        K-BERT          & -0.5016 & 0.6890 &  0.2545 & 0.5990 \\ \midrule
        After Removing  &         & \\
        \, ALIAS        & -0.4561 & 0.6731 &  0.0761 & 0.5299 \\
        \, CAT          & -0.1535 & 0.5602 &  0.3348 & 0.6292 \\
        \, DESC         & -0.2065 & 0.5806 & -0.1569 & 0.4385 \\
        \, CAT + DESC   &  0.1547 & 0.4394 & -0.2858 & 0.3891 \\
        \, ALIAS + DESC & -0.1526 & 0.5598 & -0.3317 & 0.3720 \\
        \, ALIAS + CAT  &  0.1915 & 0.4251 & -0.1911 & 0.4253 \\
        \midrule
        $\text{K-BERT}_{\text{MANUAL}}$ & -0.3911 & 0.6498 & 0.3009 & 0.6165 \\
        \bottomrule
	\end{tabular}
\end{table}

\begin{table}[ht]
    \caption{Student t-test results between test accuracies of BERT \& K-BERT Models evaluated on ag\_news\_subset dataset}
	\label{table:clsttest}
	\centering
	\begin{tabular}{l|rr|rr}
		\toprule
        \multirow{3}{*}{\textbf{Model}} & \multicolumn{4}{c}{\textbf{Knowledge-Gating}} \\
        & \multicolumn{2}{c}{\textbf{Off}} & \multicolumn{2}{c}{\textbf{On}} \\
        & \textbf{t-statistic} & \textbf{p-value} & \textbf{t-statistic} & \textbf{p-value} \\
        \bottomrule\toprule
        K-BERT          & -0.3866 & 0.3518  & -1.1488 & 0.1328 \\ \midrule
        After Removing  & & & & \\
        \, ALIAS        &  0.0100 & 0.5039  & -0.9179 & 0.1854 \\
        \, CAT          & -0.0531 & 0.4791  & -0.8354 & 0.2072 \\
        \, DESC         & -0.3869 & 0.3517  & -0.2806 & 0.3911 \\
        \, CAT + DESC   &  0.1372 & 0.5538   & 0.0739 & 0.5290 \\
        \, ALIAS + DESC &  0.1635 & 0.5640  & -0.9617 & 0.1745 \\
        \, ALIAS + CAT  & -0.3465 & 0.3665  & 0.8429 & 0.7948 \\
        \midrule
        $\text{K-BERT}_{\text{MANUAL}}$ & -8.0708 & 0.0000  & -7.0396 & 0.0000 \\
        \bottomrule
	\end{tabular}
\end{table}

\end{document}